\documentclass[11pt]{article}

\usepackage[preprint]{acl}

\usepackage{times}
\usepackage{latexsym}
\usepackage{booktabs}
\usepackage[T1]{fontenc}
\usepackage{listings}
\definecolor{keywordcolor}{HTML}{770489}

\lstdefinelanguage{SPARQL}{
  keywords={SELECT, WHERE, FILTER, OPTIONAL, PREFIX, LIMIT, DISTINCT, GROUP, ORDER, BY, DESC, COUNT},
  sensitive=true,
  morecomment=[l]{\#},
}
\definecolor{tblHeader}{RGB}{245,245,245}
\definecolor{tblRowA}{RGB}{250,250,250}
\definecolor{tblRowB}{RGB}{255,255,255}

\usepackage{lstlinebgrd}

\lstdefinestyle{freqtable}{
  basicstyle=\ttfamily\small,
  columns=fullflexible,
  keepspaces=true,
  frame=none,
  numbers=none,
  xleftmargin=0pt,
  xrightmargin=0pt,
  aboveskip=0.5em,
  belowskip=0.5em,
  linebackgroundcolor={
    \ifnum\value{lstnumber}=1
      \color{tblHeader}
    \else
      \ifodd\value{lstnumber}
        \color{tblRowA}
      \else
        \color{tblRowB}
      \fi
    \fi
  }
}

\usepackage[utf8]{inputenc}

\usepackage{microtype}

\usepackage{inconsolata}

\usepackage{graphicx}
\newcommand{\disamkblogo}{%
  \raisebox{0ex}{\includegraphics[height=1.4ex]{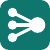}}%
}

\newcommand{\entity}[2]{%
  \href{https://gptkb.org/entity/#1}{%
    \mbox{\disamkblogo\,\texttt{#2}}%
  }%
}

\title{GPTKB 2.0: Browsing, Querying, and Auditing a Disambiguated LLM-Derived Knowledge Base}

\author{
Yujia Hu$^{1}$
\quad
Tuan-Phong Nguyen$^2$
\quad
Simon Razniewski$^1$
\\ 
\\
$^1$ScaDS.AI Dresden/Leipzig \& TU Dresden, Germany
\\
$^2$VNU University of Engineering and Technology, Hanoi, Vietnam
\\
{\small\texttt{\{yujia.hu,simon.razniewski\}@tu-dresden.de \quad tuanphong@vnu.edu.vn }}
}

\begin{document}
\maketitle
\begin{abstract}
We present a web demo for exploring a large-scale disambiguated knowledge base (KB) materialized from a large language model (LLM). GPTKB 2.0 contains 38.4M triples over 1.6M canonical entities, together with 207.6K consolidated relations and 66K consolidated classes. Unlike prior LLM-derived knowledge bases that largely identify entities by surface strings, GPTKB 2.0 performs context-guided disambiguation during recursive KB construction, separating homonyms and merging synonymous mentions as facts are elicited. The demo makes this process inspectable: users can browse entities, follow links across the KB, and audit the provenance of individual facts, including surface forms, candidate matches, source triples, and disambiguation decisions. The interface further supports structured SPARQL queries, natural-language questions translated to SPARQL, and entity linking from user-provided text to canonical GPTKB 2.0 entries. GPTKB 2.0 is available at \url{https://gptkb.org/}, with the full KB downloadable for offline use.
\end{abstract}

\section{Introduction}
General-domain KBs \cite{wikidata, dbpedia, yago} are important and longstanding backbones for AI applications, and entity disambiguation recognized as a core challenge in KB construction. More recently, large language models (LLMs) have emerged as implicit repositories of factual knowledge~\cite{petroni-etal-2019-language}, inspiring a line of work that constructs knowledge bases directly from LLMs by materializing their parametric knowledge in structured form~\cite{mango, cohen-etal-2023-crawling, gptkbv1, gptkbv1.5,parovic-etal-2025-generating}. However, these approaches largely rely on surface strings as entity identifiers. Yet surface strings are inadequate identifiers, exhibiting two complementary failure modes: without further triple context, they conflate \emph{homonymous} entities that share identical entity label (e.g., \textit{Munich} in \textit{[Europe, hasMajorCity, Munich]} vs. \textit{Munich} in \textit{[Mathieu Amalric, notableWork, Munich]}) and fragment \emph{synonymous} strings that denote the same entity (e.g., \textit{New York City} and \textit{The Big Apple}).  

\begin{figure}[t]
    \centering
    \includegraphics[
    width=\linewidth
]{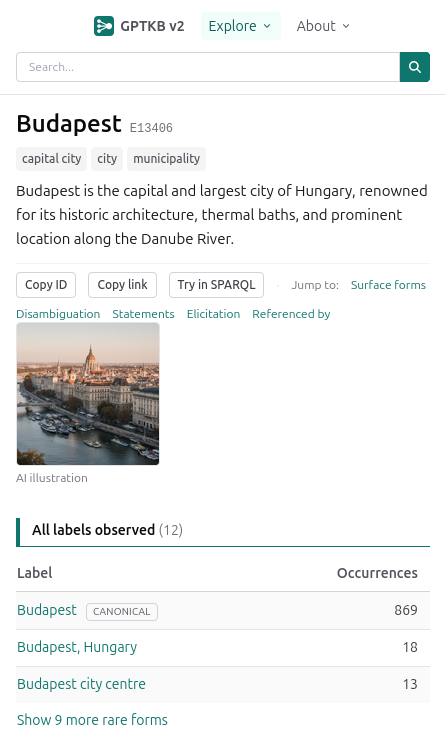}
    \caption{\entity{E13406}{Budapest} in GPTKB 2.0.}
    \label{fig:budapest}
\end{figure}

We present GPTKB 2.0, a disambiguated, general-domain KB built entirely from an
LLM, together with a web interface that makes both the knowledge and its
provenance explorable. GPTKB 2.0 comprises 38.4M triples over 1.6M canonicalized
entities, 207.6K consolidated relations, and 66K consolidated classes.
Disambiguation is performed on the fly during construction, driven by a simple
observation: the eliciting triple, together with textual descriptions of
existing entities, typically provides sufficient context to distinguish new
entities from known ones without external resources.

The demo's distinguishing feature is transparency. Through the interface, users
can (i) browse and query the full KB via SPARQL or natural language; (ii)
inspect the derivation and consolidation provenance behind any fact: all its
trajectories from the seed entity, and the surface forms, candidate matches, and
context considered during consolidation, so that both homonymy and synonymy
resolution become auditable; and (iii) link named entities in their own input
text to canonical GPTKB 2.0 entries. To our knowledge, GPTKB 2.0 is the first demo
to combine disambiguation provenance, SPARQL, and natural-language querying over
an LLM-derived KB.

\section{Construction}
GPTKB 2.0 methodology \cite{hu2026gptkb20directconstruction} represents entities, relations, and classes as first-class elements with unique identifiers, so that disambiguation can operate across homonymous and synonymous mentions. Following curated KBs such as Wikidata~\cite{wikidata}, besides unique ID, each element also carries a textual description, which serves as the context signal for disambiguation. An example for \entity{E13406}{Budapest} is shown in Figure~\ref{fig:budapest}. Starting from a seed entity, GPTKB 2.0 is built by a recursive, context-guided pipeline that expands the KB along the subject--object frontier, disambiguating each new mention on the fly. The pipeline proceeds through four context-guided steps: elicitation, named entity recognition (NER), and disambiguation.

\paragraph{Elicitation and NER}
Both steps rest on one principle: a bare label is ambiguous, so we condition the LLM on context. Given an entity, the LLM is prompted for triples; since facts about \textit{Munich} could concern the city or the 2005 film, we include the entity's description so that elicited facts pertain to the intended entity. Each elicited object is then classified as a literal or a
named entity, \textit{01099} is a literal in \textit{[Neustadt Dresden, postalCode, 01099]} but an entity in \textit{[Frisch, partOfBand, 01099]}---using the source triple as context.

\paragraph{Disambiguation} Named-entity objects are resolved against the KB to merge synonyms and separate homonyms. The LLM is prompted with the target entity label, its source triple, and the labels and descriptions of the nearest candidates by embedding similarity; it either matches an existing entity, marks it new, or abstains. A match inherits the incoming label as an alias. For instance, \entity{E21335}{Munich} (city) and \entity{E854833}{Munich} (film) are kept distinct through their differing source-triple context; \textit{The Big Apple} is mapped to \entity{E550}{New York} based on the source triple \textit{[Emerald City, contrastsWith, The Big Apple]}.

\begin{table}[t]
\centering
\small
\begin{tabular}{@{}lr@{}}
\toprule
\textbf{Entities} & 1,592,185   \\
\textbf{Triples} & 38,450,135 \\
\textbf{Relations} & 207,633 \\
\textbf{Classes} & 66,523 \\
\textbf{Triple objects (entities)} & 15.8M \\
\textbf{Triple objects (literals)} & 22.6M \\
\textbf{Avg.\ triples per entity} & 38.4 \\
\textbf{Avg.\ outlinks per entity} & 15.8 \\
\textbf{Entities novel to Wikidata}* & 36.8\% \\
\bottomrule
\multicolumn{2}{l}{\small\textbf{*} \textit{Validated on 1,000 samples.}}\\
\end{tabular}
\caption{Overall statistics of GPTKB 2.0.}
\label{tab:statistics}
\end{table}

\begin{figure*}[t]
    \centering
    \includegraphics[width=1\linewidth,
                 trim=0.5cm 3cm 0.5cm 3cm,clip
                 ]{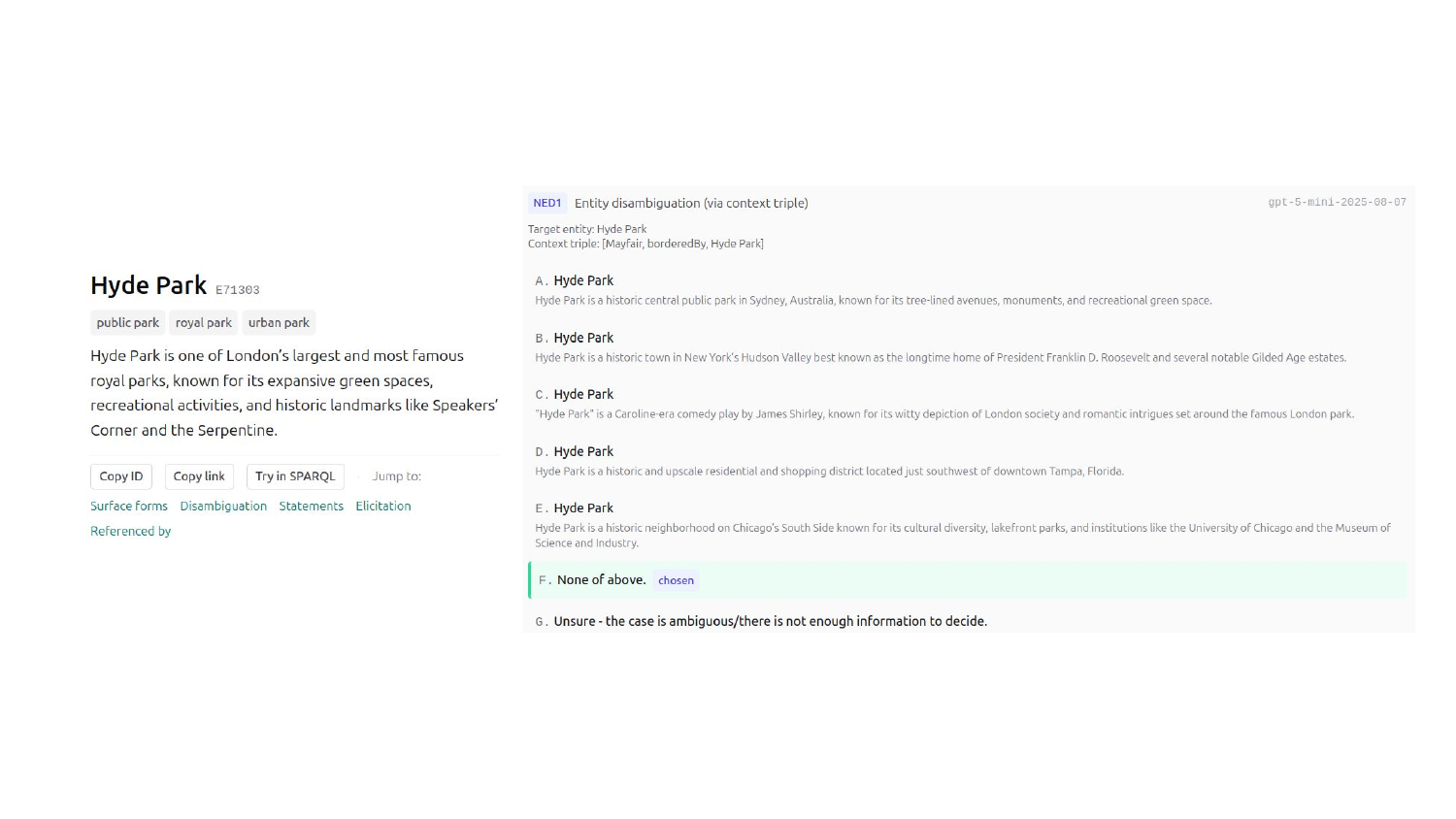}
    \caption{\textit{Hyde Park} in London is recognized as distinct from other synonymous entities.}
    \label{fig:homonymy}
\end{figure*}

\section{Overview}
As shown in Table~\ref{tab:statistics}, GPTKB 2.0 contains 38.4M triples and 1.6M disambiguated entities, together with 207.6K relations and 66.5K classes. 

\section{Web Interface}
GPTKB 2.0 is served at \url{https://gptkb.org/}, which provides a user interface for keyword-based entity search and link-based exploration to discover connected entities. The interface is built with the Django framework and served via Nginx, with the KB stored in an OpenLink Virtuoso triple store. To our knowledge, this is the first demo interface to make the consolidation process fully traceable while combining SPARQL querying, natural-language (text-to-SPARQL) querying, and entity linking over an LLM-derived KB.
\subsection{Access}
Entities can be reached through several routes:
\begin{enumerate}
    \item The start page links directly to featured entities such as
    \textit{Vannevar Bush} and \textit{San Francisco}.
    \item A search field in the top-right corner supports string-based lookup.
    \item When an entity's ID is known, its page can be accessed directly at
    \url{https://gptkb.org/entity/<ID>} (e.g., \texttt{E14} for the \entity{E14}{USA}).
\end{enumerate}


\subsection{Traceable decision-making} 
\begin{figure*}[t]
    \centering
    \includegraphics[
    width=1\linewidth,
    trim=0.5cm 2cm 0.5cm 2.5cm,
    clip
]{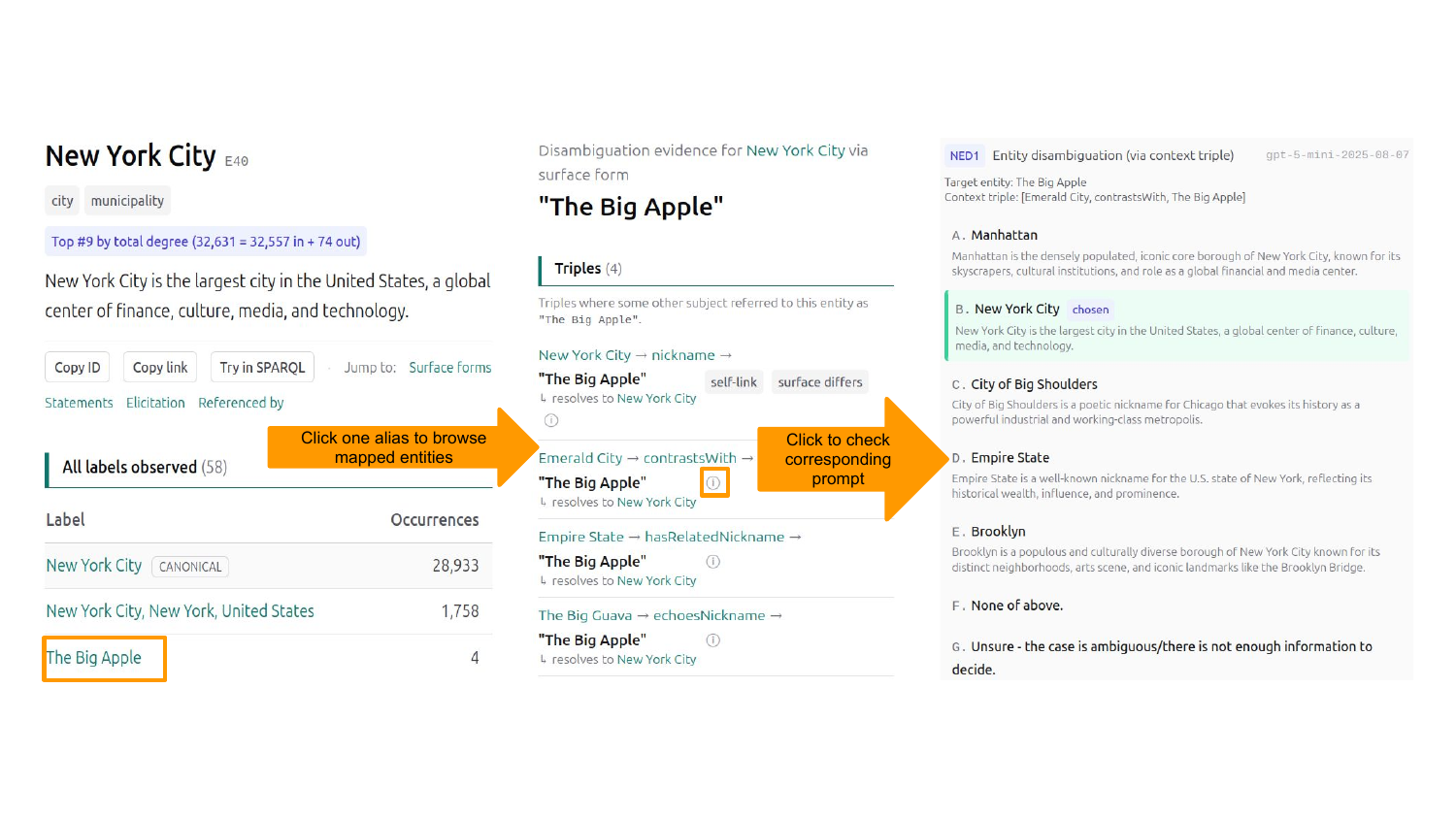}
    \caption{The Big Apple is mapped to existing entity \textit{New York City}.}
    \label{fig:synonymy}
\end{figure*}

Each entity's page exposes the disambiguation decisions behind it, making both homonymy and synonymy auditable. To trace homonymy handling, users can inspect the candidates presented to the LLM when this entity was disambiguated, revealing how it was kept distinct from same-label candidates and judged new to the KB at the point it first appeared. As shown in Figure~\ref{fig:homonymy}, clicking \textsf{Show} under \textit{How this entity was disambiguated} reveals the process by which \entity{E71303}{Hyde Park} in London is separated from its homonyms existing in GPTKB 2.0. To trace synonymy handling, the surface forms consolidated into this entity are listed as its aliases; clicking one opens the prompt for the corresponding disambiguation step, showing how that form was merged. As shown in Figure~\ref{fig:synonymy}, the alias list of \entity{E40}{New York City} lets users click an alias (here, \textit{The Big Apple}) to reveal the mentions mapped under that label; the info button then opens the prompt used at that step, listing the target entity, its source triple, and the retrieved candidates.

\subsection{Entity Linking}
Beyond showing how entities within GPTKB 2.0 were linked during construction, the
interface integrates LELA~\cite{haffoudhi2026lelallmbasedentitylinking,haffoudhi2026lelallmbasedentitylinking2}, an
entity-linking system, over GPTKB 2.0. A user enters text and clicks
\textsf{Link entities}; LELA detects named-entity mentions and links each to its
corresponding entity in GPTKB 2.0, or marks it \textsc{NIL} when no match exists.  
\includegraphics[width=\columnwidth]{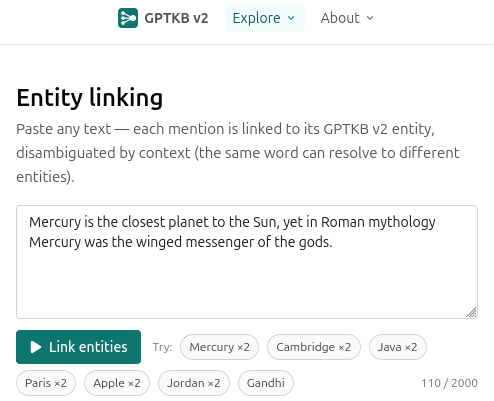}\\
Linking is also context-dependent: in the example, the two occurrences of \textit{Mercury} resolve to distinct entities, the planet (\entity{E18768}{Mercury}) and the Roman deity (\entity{E397485}{Mercury}), despite sharing a surface form.
\includegraphics[width=\columnwidth]{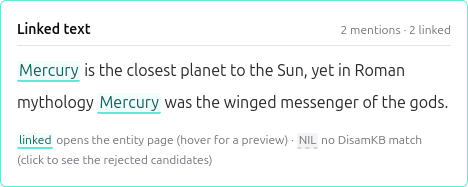}\\
Each decision is auditable through a disambiguation-evidence panel that lists every candidate the linker weighed. Candidates are retrieved in two ways: surface form matches, ranked by how often each entity has been observed under that surface form, and a BM25 text-relevance fallback that fills the remaining slots. Notably, context can override raw frequency: although the planet is the most frequent entity linked to \textit{Mercury} (observed 53 times), the mythological mention is correctly resolved to the Roman deity (observed only 16 times), which the surrounding context favors. This transparency lets users see not only which entity was chosen but which alternatives were considered and rejected.
\includegraphics[width=\columnwidth]{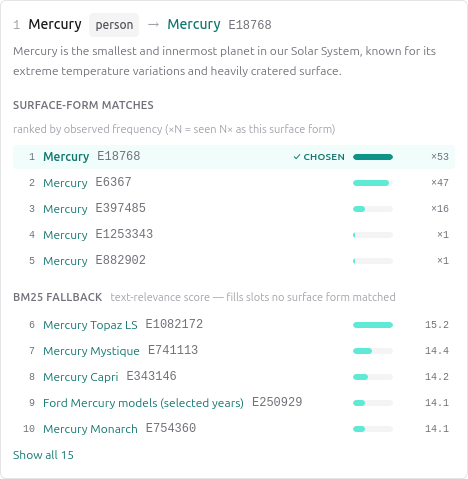}\\
\includegraphics[width=\columnwidth]{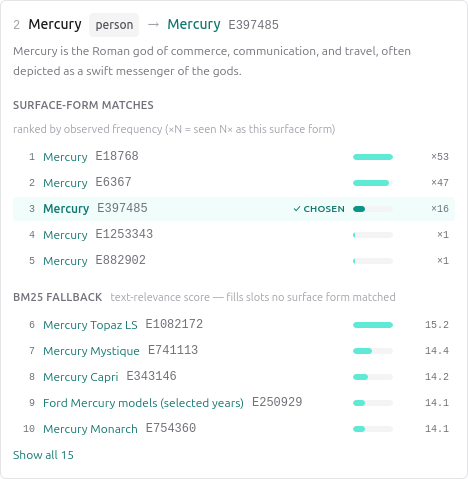}\\

\subsection{SPARQL Query}
Because GPTKB 2.0 is a fully materialized, structured KB, its contents can be queried directly rather than through repeated prompting of the underlying model. The KB is stored in an OpenLink Virtuoso triple store and exposed through a SPARQL endpoint at \url{https://gptkb.org/query/}, supporting structured queries, filtering, and aggregation over all 38.4M triples. This turns questions about the model's knowledge that would otherwise require bespoke, large-scale prompting (e.g., thousands of templated probes~\cite{Kotek_2023}) into single declarative queries evaluated with mature database technology. In the following, we show enabled analyses.
\paragraph{Most frequent classes}
The most frequent classes in GPTKB 2.0 can be queried via:
\begin{lstlisting}
PREFIX gptkbp: <https://gptkb.org/prop/>

SELECT ?o (COUNT(*) AS ?freq)
WHERE {
  ?s gptkbp:P0 ?o .
}
GROUP BY ?o
ORDER BY DESC(?freq)
LIMIT 20
\end{lstlisting}
\begin{lstlisting}[style=freqtable]
label                         freq
human                       166328
municipality                 66232
tourist attraction           51891
fictional beaver             29751
song                         25408
neighborhood                 24317
building                     23436
film                         23376
\end{lstlisting}

\paragraph{Entities with most aliases}
Entities with most aliases in GPTKB 2.0 can be queried via:
\begin{lstlisting}
PREFIX rdfs: <http://www.w3.org/2000/01/rdf-schema#>
PREFIX skos: <http://www.w3.org/2004/02/skos/core#>

SELECT ?e ?eLabel (COUNT(?form) AS ?nforms) 
WHERE { 
  ?e (rdfs:label|skos:altLabel) ?form 
} 
GROUP BY ?e ?eLabel 
ORDER BY DESC(?nforms) 
LIMIT 20
\end{lstlisting}
\includegraphics[width=\columnwidth]{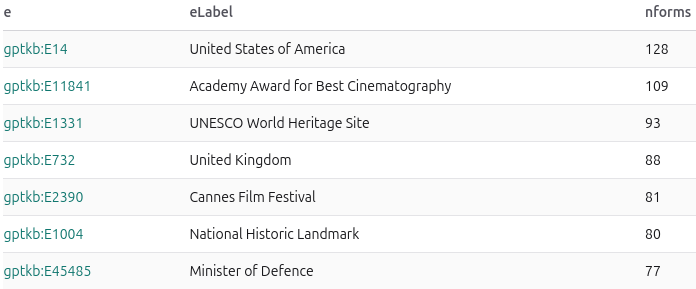}\\
As shown above, 128 distinct surface labels map to the entity \textit{United
States of America}. This illustrates the value of resolving entities during construction rather than afterward: because each merge is available to subsequent elicitation steps, later facts are gathered against the consolidated entity, avoiding the fragmentation that a post-hoc pass would first have to detect and then repair.

\subsection{Query in Natural Language}
For users unfamiliar with SPARQL, the demo integrates GRASP~\cite{DBLP:conf/semweb/WalterB25}, an agent that answers plain-English questions over GPTKB 2.0. Given a question, the agent iteratively explores GPTKB 2.0 through tool calls, resolving the entities, properties, and classes it needs, then composes and executes a SPARQL query to produce the answer. For example, \textit{``What is Budapest known for?''} is resolved by locating the \textit{Budapest} entity (\texttt{E13406}) and the \textit{notableFor}
property (\texttt{P22}), yielding a concise list of answers.
\includegraphics[width=\columnwidth]{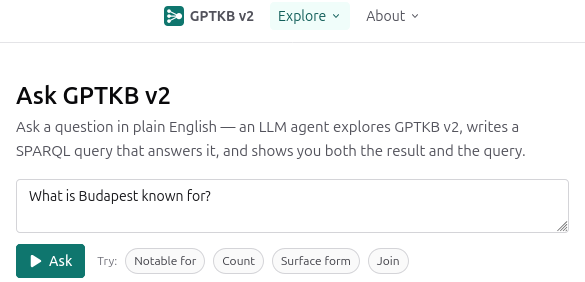}\\
\includegraphics[width=\columnwidth]{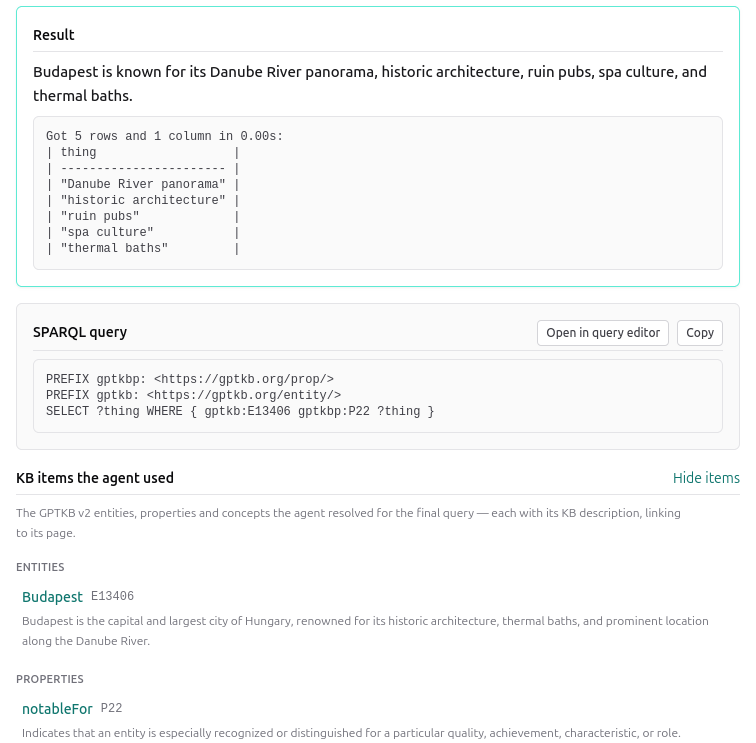}\\
Consistent with the rest of the interface, every step is transparent: the demo exposes the agent's full reasoning trace, the generated SPARQL query (which users can open in the query editor), and the KB entries the agent resolved, each linked to its
entity page.\\[5pt]
\includegraphics[width=\columnwidth]{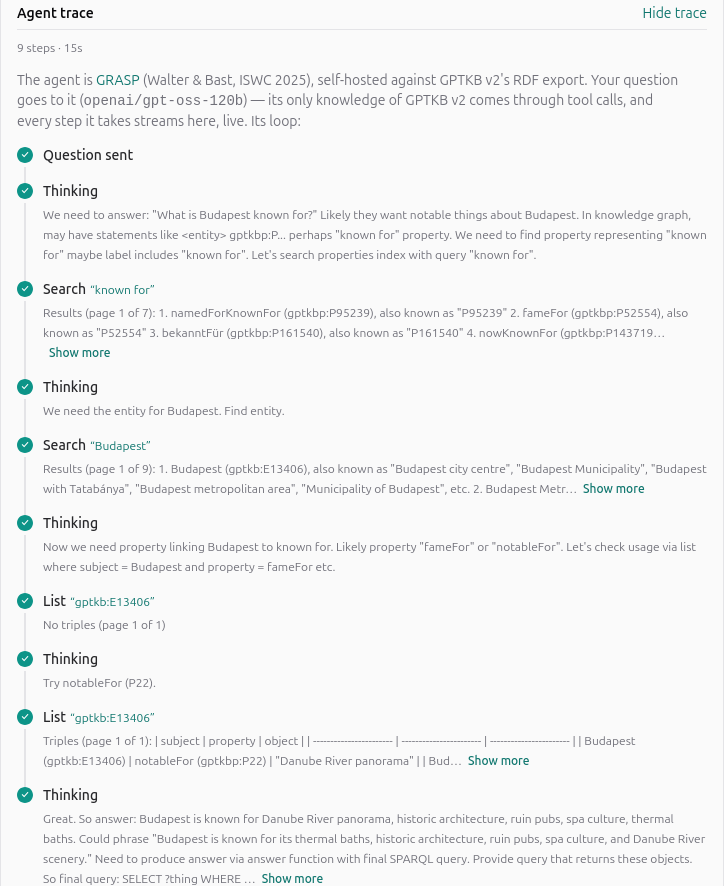}

\section{Evaluation}
\subsection{Disambiguation}
Reliable link-based exploration depends on correct disambiguation: linked
entities must be neither wrongly merged nor wrongly split. We manually assess both error directions on samples of 100 cases each
(Table~\ref{tab:evaluation}). \textbf{Merge precision} measures whether entities
combined into one are truly the same: 98\% of same-label objects merged into an
existing entity are correct (homonymy), as are 91\% of differently-labeled
aliases resolved to an existing entity (synonymy); false merges (9\%) in the
latter are the main error mode. \textbf{Split precision} measures whether
entities kept separate are truly distinct: all 100 same-label pairs assigned
distinct IDs are genuinely different (homonymy), and for 95\% of entities
no synonymous duplicate appears among their top-20 label-embedding neighbors
(synonymy; an approximation, since duplicates with dissimilar surface forms may
be missed). The pipeline thus handles both homonymy and synonymy reliably in
both directions, with false synonym merges the largest residual error.

\begin{table}
\centering
\small
\begin{tabular}{@{}lccc@{}}
\toprule
\textbf{Judge} & \textbf{Human} & \textbf{LLM} & \textbf{LLM$^-$} \\
\midrule
\multicolumn{4}{@{}l}{\textbf{NED precision}} \\
\quad Merge correct & 94.5\% & --- & ---\\
\quad Split correct & 97.5\% & --- & ---\\
\addlinespace
\multicolumn{4}{@{}l}{\textbf{Triple precision}} \\
\quad True & 94.5\% & 92.8\% & 80.0\% \\
\quad Plausible & 2.5\% & 5.5\% & 10.3\%\\
\quad Implausible & 1.0\% & 1.2\% & 2.0\%\\
\quad False \% & 2.0\% & 0.5\% & 7.7\%\\
\addlinespace
\multicolumn{4}{@{}l}{\textbf{Entity factuality}} \\
\quad Verifiable & 96.0\% & 92.3\% & --- \\
\quad Plausible & 2.0\% & 6.7\% & --- \\
\quad Unverifiable & 2.0\% & 1.0\% & --- \\
\bottomrule
\end{tabular}
\caption{Manual and automatic evaluation of GPTKB 2.0. Human evaluation uses $n=400$ for NED precision and $n=200$ for triple precision and entity factuality; LLM evaluation uses $n=1{,}000$.}
\label{tab:evaluation}
\end{table}

\subsection{Overall Quality of GPTKB 2.0}
Since exhaustive evaluation of a 38M-triple KB is infeasible, we assess randomly sampled subsets. Following YAGO~\cite{yago} and recent factuality metrics such as FactScore~\cite{min-etal-2023-factscore} and
VeriScore~\cite{song-etal-2024-veriscore}, we verify samples against
web-retrieved evidence, using two paradigms: 200 items
validated by humans and 1{,}000 items judged by an agentic pipeline.
Precision exceeds 90\% for both triples and entities
(Table~\ref{tab:evaluation}). For \textbf{triples} (labeled true, plausible,
implausible, or false), 94.5\% are true and 2.0\% false on the human sample, and
92.8\% true and 0.5\% false on the automatic sample. In an ablation where the
judge sees each triple without the subject entity's description, the true rate
drops to 80\%, underscoring how much descriptions contribute to interpretability.
For \textbf{entities} (labeled verifiable, plausible, or unverifiable), 96\% are
verifiable on the human sample and 92.3\% on the automatic sample.



\section{Related Work}
\paragraph{Disambiguated Knowledge Bases}
Managing entity identity is a longstanding challenge in KB construction. Curated
KBs (DBpedia~\cite{dbpedia}, YAGO~\cite{yago}, Wikidata~\cite{wikidata}) inherit
canonicalization from human-maintained resources, and mention-grounding methods
reuse such inventories~\cite{hoffart-etal-2011-robust}, but both bound coverage
to what those resources contain; text-based systems~\cite{Carlson2010NELL} broaden coverage yet often set consolidation aside. LLM-derived
KBs~\cite{cohen-etal-2023-crawling, gptkbv1, parovic-etal-2025-generating} are
tied to neither a fixed corpus nor a closed schema, but inherit no external
identifiers and must disambiguate from scratch.
\paragraph{Traceable KB Interfaces}
Public KBs are typically exposed through browsers, entity pages, and SPARQL
endpoints that present the KB as a finished artifact showing what it contains, not how each entry arose. For LLM-derived KBs, whose entries are generated and consolidated automatically, the provenance behind each fact and disambiguation decision is itself evidence of trustworthiness. GPTKB 2.0's interface makes this process traceable, surfacing for each fact its derivation trajectory, the candidates weighed during disambiguation, and the context behind each decision.

\section{Conclusion}
We presented GPTKB 2.0, a disambiguated knowledge base materialized entirely from LLM, together with a web interface that makes both its knowledge and the disambiguation decisions behind it explorable. The KB comprises 38.4M triples over 1.6M entities, 207.6K relations, and 66K classes, all consolidated on the fly during a recursive construction pipeline. The distinguishing feature of this demo is transparency: every fact carries provenance recording the surface forms, candidate matches, and context behind its disambiguation. Through the interface, users can browse and trace entities, query the KB in SPARQL or natural language, and link named entities in their own text against it. By turning an LLM's latent parametric knowledge into a structured, queryable, and inspectable resource, GPTKB 2.0 offers both a usable general-domain KB and a step toward making the knowledge inside LLMs open to scrutiny.




\bibliography{custom}

\appendix



\end{document}